# A review of systematic selection of clustering algorithms and their evaluation

Marc Wegmann, Domenique Zipperling, Jonas Hillenbrand and Jürgen Fleischer

*Abstract*—Data analysis plays an indispensable role for value creation in industry. Cluster analysis in this context is able to explore given datasets with little or no prior knowledge and to identify unknown patterns. As (big) data complexity increases in the dimensions volume, variety, and velocity, this becomes even more important. Many tools for cluster analysis have been developed from early on and the variety of different clustering algorithms is huge. As the selection of the right clustering procedure is crucial to the results of the data analysis, users are in need for support on their journey of extracting knowledge from raw data. Thus, the objective of this paper lies in the identification of a systematic selection logic for clustering algorithms and corresponding validation concepts. The goal is to enable potential users to choose an algorithm that fits best to their needs and the properties of their underlying data clustering problem. Moreover, users are supported in selecting the right validation concepts to make sense of the clustering results. Based on a comprehensive literature review, this paper provides assessment criteria for clustering method evaluation and validation concept selection. The criteria are applied to several common algorithms and the selection process of an algorithm is supported by the introduction of pseudocode-based routines that consider the underlying data structure.

## 1 Introduction

The ever-growing amount of data produced in society and industry alike gives rise to new applications of machine learning and data mining tasks. The volume of digital data produced each year is increasing rapidly and further exponential growth is predicted (Statista 2018). Fig. 1 shows the forecast for digital data production in 2025 compared to latest statistics. This is where statistics, computer science, and machine learning (ML) come into play. ML techniques are well suited for analyzing large amounts of data. ML generally distinguishes two types that are relevant in the context of analyzing data: supervised and unsupervised machine learning. Supervised machine learning requires a priori knowledge about patterns to be detected in the data as well as corresponding labeling. Classical tasks are regression or classification problems. Unsupervised learning on the other hand does not require any labels and searches for unknown patterns within the data, also known under its classical term pattern recognition (Russell and Norvig 2020). Classical tasks in this context are clustering problems.

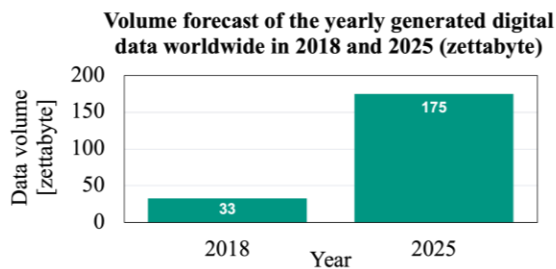

**Fig. 1 Forecast on annual digital data growth worldwide (in zettabyte) (Statista 2018)**

Clustering algorithms are applied to a variety of task with no exclusion of particular fields, as long as abundant data is available. Very prominent application fields are for example data reduction, hypothesis generation, prediction based on groups, business applications and market research, biology and bioinformatics, spatial data analysis, as well as web mining (Gunopulos 2009).

The main goal of every machine learning approach in general and data clustering in particular is to extract knowledge from data, but as stated by (Xu and Wunsch 2005): The selection and design of a suitable clustering algorithm is a crucial journey. There are three major challenges for novel users applying cluster analysis:
- A general challenge of machine learning poses the „No free lunch theorem". (Wolpert 1996) states that a priori distinctions between learning algorithms are not possible in noise-free scenarios (Sewell and Martin).
- There is no appropriate central repository for cluster technique evaluation.
- Cluster evaluation metrics must be aligned with the choice of cluster algorithms and also depend on the underlying data distribution. Especially, internal metrics are by definition unsupervised (see chapter 2.2.3).

Hence, a-priori decision which cluster algorithm and evaluation to choose, is not a simple task, especially for novices in the field of machine learning. The main goal of this paper is to provide a comprehensive analysis and overview of state of the art clustering techniques. We want to present a categorization system and selection tool for cluster algorithms and furthermore corresponding clustering metrics to evaluate the clustering results.

Therefore, the remaining chapters of this paper are organized as follows. In "Preliminaries" relevant terms are defined to provide a common terminology for the context of this work. Chapter two "State of the art: *Clustering techniques*" introduces state of the art clustering algorithms and cluster evaluation. The section concludes with a summary and current deficits in both topics. In the third chapter we propose a systematic selection procedure for both cluster algorithms and cluster evaluation, incorporating the state of art and research. Finally, we introduce some pseudocode-based routines to support the selection process in chapter four. The approach is then discussed and we close the paper with a conclusion and an outlook.

## 1.1 PRELIMINARIES

In the following, we define or specify terminology used in the context of this paper.

**Definition: Clustering**

First of all a common understanding of the term "Clustering" is required. (Fung 2001) emphasizes that a large number of clustering definitions can be found in literature. The simplest definition is shared among all and describes the central concept of clustering as grouping of similar data items. Similarity in this context can be determined in different ways, for example based on mathematical distance measures. Although a common agreement on the term of clustering is not found yet and will probably never be found in the future, a more elaborate definitions is given by (Jain and Dubes 1988) who define clustering based on three properties:
(1) Instances in the same cluster must be similar as much as possible to each other
(2) Instances in the different clusters must be different as much as possible to each other
(3) Measurement for similarity and dissimilarity must be clear and have a practical meaning

**Definition: Taxonomy**

A system for naming and organizing things and concepts into groups that share similar qualities is referred to as a "taxonomy" (Cambridge Dictionary 2021). In the context of this paper, a taxonomy is used to categorize clustering algorithms into groups based on their underlying logic. A taxonomy for clustering algorithms is presented later in this work.

**Definition: Cluster Analysis**

According to (Xu and Wunsch 2005), a typical cluster analysis consists of four central steps with a feedback pathway. Given some input data, the first step is the feature selection or extraction. In feature selection, distinguishing features are selected from several candidates. In feature extraction on the other hand, some transformations of original features are used to create meaningful and new features. Both are very important for the effectiveness of clustering applications and an elegant feature selection/extraction can significantly reduce the workload and simplify the subsequent design process. In general, ideal features should be immune to noise as well as easy to extract and interpret.

The second step is the design or selection of a capable clustering algorithm. The selection of a clustering algorithm directly affects the clustering results. Since the number of possible solutions is very rich in literature, it is very important to carefully investigate the characteristics of the underlying problem to select or design an appropriate approach.

Subsequently, the application of the chosen algorithm needs to be validated. Given a certain dataset, a clustering algorithm will always find a division. The key is to divide the data in a way that serves the solution of the initial problem. Therefore, effective evaluation criteria for the clustering results are needed. In general, three concepts can be distinguished in this context: external, internal, and relative indices (Halkidi and Vazirgiannis 2008). The choice of the right evaluation criteria is also crucial to the evaluation results and highly depends on the underlying data as well as the applied clustering algorithm.

In a last step, the clustering results need to be interpreted. The ultimate goal here is to extract meaningful insights from the original data that solve the initial data clustering problem. Fig. 2 summarizes the general clustering procedure. As highlighted in the figure, the focus of this paper is on providing a tool for the clustering algorithm selection and validation step.

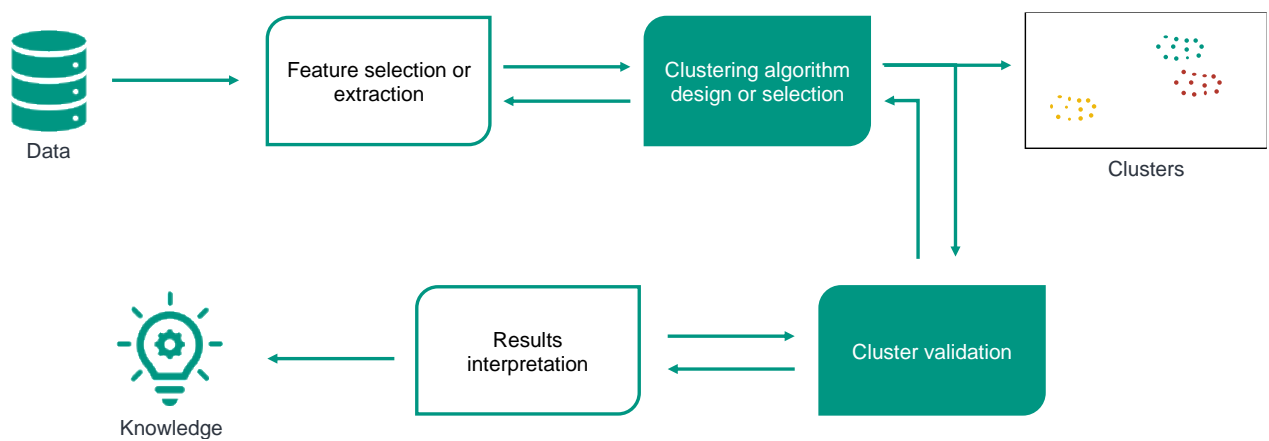

**Fig. 2 Procedure of the cluster analysis process**

**Definition: Cluster Validation *Metric* and *Index***

To avoid confusion throughout this paper, the term metric or validation metric is always used to refer to some superficial concept of validation, while the term index only refers to actual proposed or applied validation methods such as the $S_{Dbw}$-Index.

## 2 STATE OF THE ART: CLUSTERING TECHNIQUES

As already stated above, the clustering algorithm selection and the cluster validation process both are crucial in the process of knowledge generation from data. This chapter gives an overview of existing literature in those two fields. Therefore, clustering algorithms and existing work regarding their applicability are presented in chapter 2.1. Cluster validation concepts on the other hand are presented in chapter 2.2. Finally, insights from the state of the art and corresponding research currents are derived in chapter 2.3.

### 2.1 CLUSTERING ALGORITHMS AND TAXONOMY

The amount of different clustering algorithms existing today is huge. The most popular among them is probably the k-means algorithm by MacQueen in 1967. From those early years on, more and more clustering algorithms for different purposes were further developed. Driven by big data and increasing computational power, novel clustering algorithms are still developed and presented today to answer current business needs.

In order to structure those clustering algorithms, a suitable taxonomy is needed. A widely used approach in literature is to distinguish between partitioning-based, hierarchical, density-, grid-, and model-based clustering algorithms. In order to also represent current research in the field of clustering algorithms that does not fit this traditional categorization, the group of novel approaches is added at this point. The taxonomy used in this paper is described in more detail below. Fig. 3 gives an overview and shows exemplary algorithms for each group.

**Partitioning-based algorithms** divide a database D of n objects into a set of k partitions, where each partition k represents a certain cluster. Corresponding algorithms typically start with an initial partition of D. In this first step, each partition (or cluster) is represented by the gravity center of the cluster (k-means algorithms) or by one of the objects of the cluster located near its center (k-medoid algorithms). Partitioning-based algorithms then iteratively optimize an objective function in order to improve the quality of the initial partition. In a last step, each of the n objects is assigned to the cluster with its representative closest to the considered object. The assignment of data objects to clusters implies that a partition is equivalent to a Voronoi diagram, where each Voronoi cell represents a certain cluster. Examples for partitioning-based clustering algorithms are k-means, PAM, CLARANCE, or the fuzzy c-means algorithm (Fahad et al. 2014; Xu et al. 1998).

**Hierarchical algorithms** organize the database D in a hierarchical manner according to a proximity matrix. In general, the hierarchical decomposition is represented by a dendrogram, which is a tree with the whole database as a root that iteratively splits D into smaller subsets until each leaf node represents one single data object. The clustering itself can eighter be executed agglomerative (bottom-up) or divisive (top-down). Agglomerative algorithms start with the leaf nodes of the tree and merge two or more appropriate clusters until a certain termination condition is fulfilled. Divisive algorithms on the other hand start with the whole dataset as one single cluster and iteratively split the cluster in two or more most appropriate sub clusters, again until a certain termination criterion holds. In contrast to partitioning algorithms, hierarchical algorithms in general do not need the number of clusters k as an explicit input parameter. Nevertheless, a termination condition goes hand in hand with the number of resulting clusters. BIRCH, CURE, and ROCK are commonly used representatives for hierarchical clustering algorithms (Fahad et al. 2014; Xu et al. 1998).

The basic idea of **density-based algorithms** is that data objects are partitioned based on regions of high density. Each region of high density represents a data cluster. Corresponding algorithms iteratively scan every data object (or some representatives) and classify them into core points, (directly) reachable points, and outliers based on their neighborhood. Data points are then assigned to a common cluster if they are "density-reachable". "Density-reachable" in this context means that there exists a chain of core points between two data objects that are either core points itself or reachable points. A cluster can thus grow in any direction that density leads to. Therefore, density-based algorithms are in general capable of discovering clusters of arbitrary shapes and are able to detect outliers and noise. DBSCAN, OPTICS, and DENCLUE are algorithms which are exemplary for this class (Fahad et al. 2014; Xu and Tian 2015).

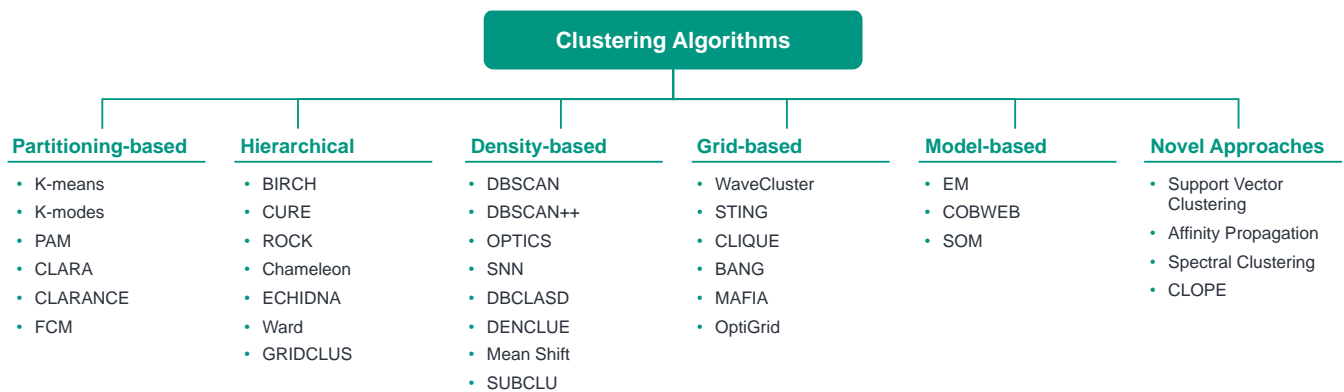

**Fig. 3 Taxonomy for clustering algorithms**

For **grid-based clustering algorithms**, the original data space is divided into a grid structure of defined size in a first step. For each grid, statistical values are calculated which are then used for the actual clustering process. The clustering process can differ but in general grids in regions of high dense are assigned to a common cluster. The accumulated grid-structure make grid-based clustering algorithms independent of the number of data objects and can result in a fast processing time, since corresponding algorithms perform the clustering on the statistical information of the grids instead of the whole dataset. The performance of a grid-based technique then mainly depends on the size of the grid, which is usually much smaller than the whole dataset. WaveCluster, STING, and CLIQUE are typical examples for grid-based clustering approaches (Fahad et al. 2014; Xu and Tian 2015).

**Model-based algorithms** optimize the fit between the given database and a particular model for each cluster. The basic idea behind such algorithms is that data is generated by a mixture of underlying probability distributions. There are two kinds of model-based clustering algorithms: one based on statistical learning and one based on neural network learning. Statistical approaches make use of probability measures in determining the clusters. Neural network approaches use a set of interconnected neurons, where each connection has an associated weight. Model-based algorithms lead to advantages such as automatically determining the number of clusters or handling noise and outliers. COBWEB or self-organizing maps are popular examples of model-based algorithms (Fahad et al. 2014).

The category **"Novel approaches"** represents clustering algorithms from recent years that do not fit the classical clustering categories described above. Due to their unique procedures, such as affinity propagation or the usage of support vectors, an assignment to existing classes is not possible. It is also not goal oriented to define a unique class for each of these algorithms on the other hand. So, grouping them in a common class is the most suitable procedure.

*2.1.1 Comparative Works on Clustering Algorithms*

Besides the above mentioned, widely used algorithms, there are numerous others that are used for more specific tasks. Due to this large scale, many authors tried to define fields of application that are associated with the algorithms itself and performed comparative studies. An overview of the state of the art is given in the following.

**Fisher & van Ness (1971)**: The two authors formally analyzed clustering algorithms with the aim of comparing them and providing guidance in selecting the right clustering procedure for a given problem. They stated that it is impossible to determine a "best" clustering algorithm but emphasized the importance of a procedure that suggests clustering algorithms that are at least admissible. The procedure they proposed consists of nine admissibility conditions which are all applied to several standard clustering methods of their time. The conditions are based on the way clusters are formed, the structure of the data, and sensitivity of the clustering technique to changes that do not affect the structure of the data. To give a more concrete example, admissibility conditions used by Fisher & van Ness are for instance:

- *Convex:* A clustering algorithm is convex-admissible if the convex hulls of clusters do not intersect in the clustering result.
- *Cluster proportion:* A clustering algorithm is cluster proportion-admissible if the cluster boundaries do not change even if some of the clusters are duplicated an arbitrary number of times.
- *Monotone:* A clustering algorithm is monotone-admissible if the clustering results do not change when a monotone transformation is applied to the elements of the similarity matrix.

They came to the result that it is impossible to create a clustering algorithm that serves all the defined conditions (Fisher and van Ness 1971).

**Dubes & Jain (1976)**: Dubes & Jain set some guidelines for a potential user of a clustering technique. For doing so, they examined eight clustering programs which are representative of the various techniques available at that time and compared their performances from several points of view. The evaluation criteria were similar to the ones used by (Fisher and van Ness 1971) and a formal comparative analysis was added with a portion of "Munson's handprinted character" dataset (Dubes and Jain 1976).

**Jain et al. (1999)**: Jain, Murty, and Flynn presented a holistic overview of clustering methods with the goal of providing useful advice regarding their application. Regarding the assessment criteria for evaluating the applicability of certain algorithms, the authors referred to the admissibility criteria of (Fisher and van Ness 1971) and exposed different shortcomings in the previous works. They criticized the lack of assessment dimensions that focus on (1) the integration of domain knowledge in the decision-making process, (2) the performance of the application to large datasets, and (3) the possibility of running an algorithm incrementally. These three issues motivated Jain et al. to evaluate several algorithms exactly based on these three new evaluation criteria. The taxonomy used in the authors' work separates clustering techniques in hierarchical and partitional approaches, which are then further divided into subcategories (Jain et al. 1999).

**Kleinberg (2002)**: Kleinberg addressed a similar problem, as the papers presented above. For the evaluation of clustering algorithms, the author defined three criteria:

- *Scale invariance:* An arbitrary scaling of the similarity metric must not change the clustering results.
- *Richness:* The clustering algorithm must be able to achieve all possible partitions on the data.
- *Consistency:* By shrinking within-cluster distances and stretching between-cluster distances, the clustering results must not change.

Beside the similar procedure, Kleinberg also provides results similar to that of (Fisher and van Ness 1971) showing that it is impossible to construct an algorithm that satisfies all of these criteria, as the title of his paper already indicates ("An Impossibility Theorem for Clustering"). Against this, further discussions in his paper expose that a "good" clustering algorithm can indeed be designed by softening the definition of satisfying a condition from "totally satisfied" to "nearly satisfied" (Kleinberg 2002).

**Liao (2005)**: Liao investigated the application of clustering algorithms to time series data. The author surveyed different studies on this subject and organized them into three groups depending on whether they work directly with the raw data, indirectly with features extracted from the raw data, or indirectly with models built from the raw data (Liao 2005).

**Xu & Wunsch (2005)**: Xu and Wunsch highlighted the importance of clustering algorithm design and selection in the overall clustering procedure. They surveyed clustering algorithms appearing in statistics, computer science, and machine learning, and illustrated their application in some benchmark datasets such as the travelling salesman problem. As a result, Xu & Wunsch defined some properties that are important in general regarding the efficiency and effectiveness of novel algorithm. These properties include for example the cluster shape, the handling of noise, and the immunity to the effects of the order of input patterns (Xu and Wunsch 2005).

**Berkhin (2006)**: Berkhin provided a comprehensive review of different clustering techniques in data mining. The author defined a list of properties of clustering algorithms that can serve as assessment dimensions in a holistic algorithm comparison. Among them are for example the attributes an algorithm can handle, the scalability to large datasets, and the dependency on data order. These properties are pretty similar to the ones defined by (Xu and Wunsch 2005). The regarded algorithms are not always assessed with respect to all of the defined properties in the study, but only to a few selected ones (Berkhin 2006).

**Ilango & Mohan (2010)**: Ilango & Mohan surveyed existing grid-based clustering algorithms and compared their effectiveness in clustering objects. For doing so, they examined the complexity of the algorithms and their input parameters. Moreover, they stated some advantages and disadvantages of the single procedures compared to each other (Ilango and Mohan 2010).

**Silva et al. (2013)**: The authors surveyed clustering techniques with the objective of clustering data streams. For this purpose, they defined some restrictions for successful procedures. From these restrictions, special requirements were derived, for example:

- Performing fast and incremental processing of data objects
- Scale to the number of objects that are continuously arriving
- Detect the presence of outliers and act accordingly

Existing traditional clustering techniques from literature are reviewed based on the fulfillment of these restrictions and thus in regard to their applicability to data streams (Silva et al. 2013).

**Fahad et al. (2014)**: Fahad et al. analyzed clustering algorithms based on the applicability to big data. For this purpose the authors defined some criteria to benchmark different clustering algorithms, which are based on the three-dimensional properties of big data: Volume, velocity, and variety. Examples for assessment criteria of the volume dimension are:

- The size of the dataset
- The handling of high-dimensional data
- The handling of noise

After a theoretical comparison with respect to the big data properties, the most promising algorithms from each algorithm class were empirically evaluated through the application on real-world traffic datasets. The authors concluded that none of the clustering algorithms perfectly matches all the properties and also do not perform great in all of the datasets (Fahad et al. 2014).

**Xu & Tian (2015)**: Xu & Tian highlighted that each clustering algorithm has its own strengths and weaknesses due to the complexity of information. They reviewed traditional as well as modern algorithms comprehensively and compared them to each other. Overall, 26 traditional algorithms from nine categories and 45 modern algorithms from ten categories were examined. Assessment criteria used were for example the time complexity, the scalability, or the applicability to large and high-dimensional data. The results give readers a systematical and clear view of data clustering methodologies (Xu and Tian 2015).

**Sajana et al. (2016) & Nayyar & Puri (2017):** These two papers are presented together because both of them extended the work of (Fahad et al. 2014). While Sajana et al. applied the same procedure to a larger range of algorithms, Nayar & Puri described related works regarding the algorithms' assessment in more detail (Nayyar and Puri 2017; Sajana et al. 2016).

**Saxena et al. (2017):** The authors give an overview of existing clustering methods and developments made at various times. They discuss characteristics and limitations of twelve clustering algorithms, among them classical representatives of the above-mentioned algorithm categories such as k-means, BIRCH, and DBSCAN. Furthermore, several evaluation criteria for the observed algorithms are defined and assessed, focusing on scale and dimensionality of the data as well as their sensitivity to outliers. In the end, the application of clustering in image segmentation, object and character recognition, information retrieval as well as data mining are highlighted (Saxena et al. 2017).

## 2.2 CLUSTER VALIDATION CONCEPTS

In this chapter we introduce the idea of cluster validation and explain its challenges by use of an example.

### 2.2.1 Purpose of cluster validation

(Halkidi and Vazirgiannis 2008; Deborah et al. 2010) state that results of clustering algorithms highly depend on user-defined input parameters and random initialization. A good example is the well-known k-means algorithm: Here, the value for *k* and the random initialization of the centroids effect the clustering results strongly.

To select appropriate and in the best-case optimal values for the input parameters as well as compare different partitions of the same data set, standardized metrics for analyzation are necessary. Thus, the main goal of using validation indices is to find the optimal partitioning of a dataset and furthermore the input parameters leading to those results. The term optimal refers not only to the correct number of clusters, but also to the correct assignment of data points to their respective clusters. The term optimal or best must be understood in the contexts of the used validation index, since every index assumes different concepts and properties of preferable clusters in order to determine characteristics of a good and bad partition.

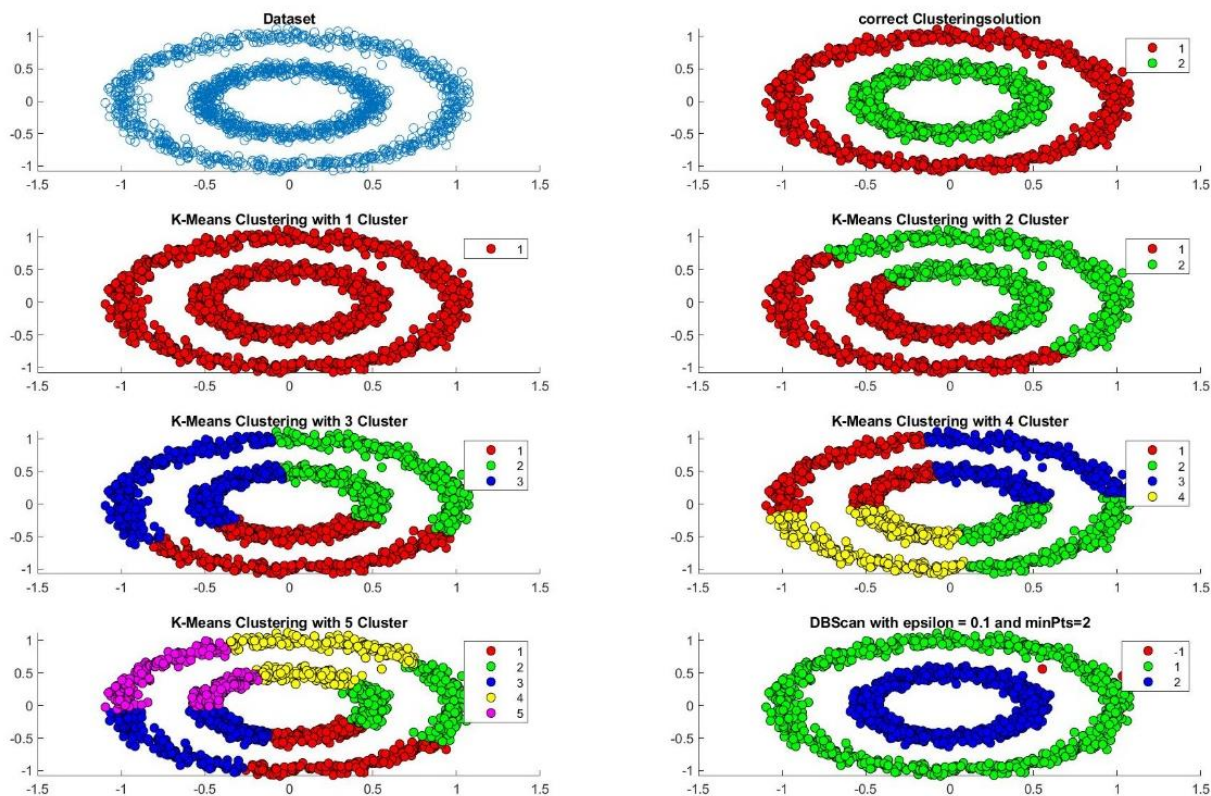

**Fig. 4 Different partitioning of the dataset based on algorithm and parameters**

*2.2.2   Connection of algorithm and validation methods*
(Handl et al. 2005) state that the selection of an internal validation index might not be independent of the used clustering algorithms since both are based on certain assumptions about the cluster structure. This results in a bias of an index towards a specific algorithm. To select a combination of algorithm and index that permits drawing meaningful conclusions, it is essential to comprehend the working principles of the algorithms and the evaluation indices used. Furthermore, it is stated that a good partition tends to perform reasonably well under multiple indices. If it performs only under one of them it might indicates that there is a bias towards the employed algorithm.

**Table 1 Index values for partitions of exemplary dataset**

| Description | Silhouette-Score | SDbw-Index | CDbw-Index |
|---|---|---|---|
| "correct clustering" | 0.11398 | 1.4907 | 0.10803 |
| "K-Means 1 Cluster" | NaN | NaN | NaN |
| "K-Means 2 Cluster" | 0.35366 | 0.64632 | 0.056465 |
| "K-Means 3 Cluster" | 0.38526 | 0.77737 | 0.016514 |
| "K-Means 4 Cluster" | 0.3763 | 0.59222 | 0.02166 |
| "K-Means 5 Cluster" | 0.35921 | 0.45949 | 0.019488 |
| "DB-Scan Cluster" | 0.11441 | 1.4989 | 0.11528 |

For illustration, we applied the k-means algorithm for values of $k = 1, \ldots, 5$ as well as the DBSCAN algorithm[1] to the two-dimensional synthetic data set shown in Fig. 4. Afterwards, three different validation indices are applied to each partition of the data set. The used indices are the average Silhouette Score ($SilSc$), the $CDbw$-Index, where high numbers indicate preferable partitions, and the $S_{Dbw}$-Index, where small numbers indicate preferable clusters. The $SilSc$ is constructed for proximities through ratio scale e.g. the euclidean distance (Rousseeuw 1987). The $S_{Dbw}$- Index tries to combine different properties such as compactness and separation of clusters to determine a score. The measurement of those properties is also based on the euclidean distance. The index tries to take the concept of density into account as well (Vazirgiannis and Halkidi 2001). The $CDbw$- Index evaluates the clustering solution through properties like compactness and separation with respect to a concept of density. The computation is based on the selection of representative points of each cluster which also tries to account for the shape of the clusters (Halkidi and Vazirgiannis 2008).

The values of the indices for the different solutions are shown in Table 1. The - not all encompassing - example clearly shows that based on the $SilSc$ and the $S_{Dbw}$-Index all solutions derived from the k-means algorithms should be considered better, while the ground truth[2] leads to lower values and hence is considered worst. This demonstrates that selecting a suitable validation index is not trivial and should be done with care. To select an appropriate index, different dimension should be considered, such as scalability, cluster shape bias, cluster number bias, complexity. This is especially important when working with high-dimensional data since a visual evaluation of the partition is not possible.

---

[1] Parameters were predefined to get a good solution, $\epsilon = 0.1, \ minPts = 2$
[2] Is only provided for comparison

*2.2.3    Categorization of cluster validation metrics*
(Halkidi and Vazirgiannis 2008) summarized the different types of validation metrics as followed: The first group of metrics are called external metrics which try to compare different partitions based on predefined class structures as a ground truth. The second group is called internal metrics which contains the group of the relative metrics. Internal metrics use found clusters and the underlying data to make a judgement about the quality of the respective solution. The relative metrics compare results for different input parameters and answer the question, which solution is preferable. In other words, they try to rank different solutions. Since clustering is an unsupervised task without a-priori information, this paper's scope mainly focuses on internal metrics.

Every metric, independent of its category, tries to quantify the quality of a clustering solution based on different properties and assumption about what constitutes a good partition of the data. Although there is no universal way of how a suitable and good metric should be build, the literature proposes different restrictions for suitable metrics.

(Amigó et al. 2009) propose four different constraints for a suitable external metric to follow. The first one is homogeneity, which essentially means that a partition $C_1$ of a data set containing a cluster, that contains all objects of two different classes $K_i$ and $K_j$, should be considered less optimal than a partition $C_2$ - of the same data set - which is identical to the first one except that all objects of class $K_i$ and $K_j$ are now in respective clusters. The second constraint is completeness which means that objects of the same class should belong to the same cluster. A metric that follows that constraint should favor a partition $C_1$ in which all objects of class $K_i$ are in the same cluster over an identical partition where the objects of class $K_i$ are in two or more different clusters. These are the most basic restrictions a metric should follow. The third constraint is referred to as rag-bad and is based on the concept, that "introducing disorder to a disordered cluster is less harmful than introducing disorder into a clean cluster" (Amigó et al. 2009). The last constraint is cluster size versus quantity which essentially boils down to the idea that "a small error in a big cluster should be preferable to a large number of small errors in small clusters" (Amigó et al. 2009).

Internal metrics try to measure the quality of a clustering solution based on the found clusters and the intrinsic information of the underlying data. (Handl et al. 2005) divides the internal metrics in three to four groups of validation techniques based on the measured properties. The first group measures the compactness of the cluster e.g. based on the intra-cluster variance. In general, there are different approaches to measure the compactness of clusters which will not be within the scope of this paper. The second group tries to assess the connectedness or more precise how well a partition agrees with the concept to what degree items are grouped together with their nearest neighbor. The third group contains all measures quantifying the cluster separation degree (Amigó et al. 2009). One example is the weighted inter-cluster distance. The fourth group contains metrics that are combination of the first three and thereby represent a group of more advanced approaches. Those metrics compute the independent score of compactness and/or connectedness and/or separation to a final score. Through this it is possible to account for trade-offs between different concepts. E.g. the intra-cluster compactness improves with a higher number of clusters while the distance between the clusters and therefor the separation worsens. (Deborah et al. 2010) take the concepts of compactness and separation and expand those with additional concepts – exclusiveness and incorrectness[3] and derive an optimizable general function which a suitable metric might follow. Exclusiveness is measured based on a probability density function and is supposed to find irregularities within the data and identify outlier. Incorrectness tries to estimate the loss of the quality of a wrongly assigned object. The verbal function is defined as followed:

$$OBF = Min(compactness) + Max(separation) \\ + Max(exclusiveness) \\ + Min(incorrectness)$$

## 2.3    INSIGHTS FROM THE STATE OF THE ART

A lot of work was put in the comparison of different clustering algorithms to each other from early on in literature. Early works, such as the papers by (Fisher and van Ness 1971) and (Dubes and Jain 1976), gained the important insight that there is no perfect clustering algorithm that serves all subjects, moreover there is always a tradeoff between different assessment dimensions depending on the underlying data clustering problem. This is also confirmed by later works, for example the one by (Kleinberg 2002). Unfortunately, these early works do not match the demands made on clustering algorithms nowadays. Important aspects such as the applicability to large datasets are missing for example.

Such requirements are then defined in later works which also introduce assessment dimensions for modern clustering algorithm such as addressing the high-dimensionality or the scalability to large datasets (Jain, Murty & Flynn 1999; Xu & Wunsch 2005; Berkhin 2006). Although these works move in the right direction of presenting a holistic algorithm comparison, the algorithm analysis is incomplete and for each algorithm only specific assessment dimensions and application fields are picked.

The most important works regarding the assessment of modern clustering algorithms are the ones of (Fahad et al. 2014), as well as the one of (Xu and Tian 2015). In both papers, assessment criteria serve the application to modern algorithms and are clearly defined in advance before being applied to a large amount of clustering algorithms from recent literature. While Fahad et al. were completely focusing on the application to big data (and leaved other aspects out of the focus), Xu & Tian compare the algorithms in a more general way. Nevertheless, the assessment dimensions and also the results are pretty similar and can serve as a good basis. Still, both papers have some shortcomings:

- *Domain information is not addressed in the papers:* It is for example of high importance, what input parameters a user has to think of before applying a certain algorithm. The choice of the input parameters highly affects the outcome of the algorithm. The most important question arising from the input parameters in this context is probably if the number of clusters has to be determined in advance. While for the k-means algorithm the number is fixed a priori, the number of clusters are only indirectly controllable for DBSCAN.
- *The output format is not addressed in the papers:* Regarding the interpretability of a clustering result, it is of high importance to keep the output structure in mind.
- *The implementation aspect is not addressed in the papers:* The pure existence of a well-fitting clustering algorithm to a given problem of a user does not solve the problem. There also need to be an implementation in a widely used programming language available, which is not necessarily given for all regarded algorithms.

---

[3] Based on statistics and estimation theory

By bringing the work of (Fahad et al. 2014) and (Xu and Tian 2015) together as well as taking care of the existing shortcomings, a holistic selection logic for modern clustering algorithms can be elaborated. Explicit requirements regarding this procedure are pictured in chapter three. As demonstrated above, after the selection of a suitable algorithm there is need to evaluate the found partitions to quantify their quality and compare different solutions for different input parameter with the final goal to find a partition that fits the underlying structure of the data or in other terms to approximate the underlying concept as well as possible. It is also clear that the creation of validation indices as well as the decision which one to use is not trivial and has to be executed with care. During this paper we will categorize some proposed indices and develop a selection tool for said metrics to set a foundation for a fast but also thorough selection in chapter three.

## 3 PROPOSAL OF A SYSTEMATIC SELECTION LOGIC FOR CLUSTERING ALGORITHMS AND VALIDATION CONCEPTS

The main goal of every machine learning approach in general and data clustering in particular is to extract knowledge from data. (Xu and Wunsch 2005) stated that the selection and design of a suitable clustering algorithm as well as finding a matching concept for validation is crucial on this journey. Depending on the characteristics of the underlying data and business problem, the performance of a certain algorithm varies. These characteristics need to be identified and evaluated in regard to the applicability of the most important clustering algorithms used in recent practice. The overall goal statement of this chapter thus addresses the algorithm selection and the cluster validation step of clustering approaches and lies in the identification of a systematic logic for the selection of suitable clustering algorithms and a corresponding validation concept for a given dataset and business problem. It is important to say, that there is no way to find the "best" algorithm or validation metric, but this chapter aims for finding admissible and in general well-performing representatives. From this goal statement, two central research questions (RQ) can be derived:

- (RQ1) Which assessment dimensions can be used in order to assess the applicability of a clustering algorithm to a certain dataset/business problem?
- (RQ2) Which assessment dimensions can be used in order to assess the applicability of a clustering validation metric?

RQ1 defines some requirements regarding the assessment dimensions in advance:

(i) The assessment dimensions must be able to evaluate state of the art clustering algorithms that are applied to current data clustering problems. These are mainly marked by big data characteristics.
(ii) The assessment dimensions must serve an application-oriented approach. Each dimension must be evaluated unambiguously, and it must be possible to extract a systematic selection logic from the evaluation results.

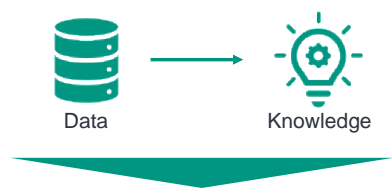

Systematic logic for the selection of suitable clustering algorithms and a corresponding validation concept for a given dataset and business problem

**Research Question 1**

Which **assessment dimensions** can be used in order to assess the applicability of a clustering algorithm to a certain dataset/business problem?

**Research Question 2**

Which **assessment dimensions** can be used in order to assess the applicability of a clustering validation metric?

**Fig. 5 Goal statement and corresponding research question**

RQ2 defines dimensions in order to categorize different validation indices to assess their applicability to a certain validation problem. Fig. 5 summarizes the goal statement and corresponding research questions of this chapter.

## 3.1 ASSESSING CLUSTERING TECHNIQUES

In order to fulfill RQ1, twelve assessment dimensions are introduced based on literature review and own assumptions. They can be grouped into five categories. The categories volume, velocity, and variety focus on the applicability to big data. The categories hyperparameters & output as well as application focus on general and application-oriented features of clustering algorithms. Each of the twelve assessment dimensions are explained in detail below and a summary is presented in Fig. 6.

**Inputs**

The nature and the amount of input parameters highly affect the quality of the clustering results. The more input parameters an algorithm needs and the more complex it is to find good ones, the less practical an algorithm is in the end. The desirable feature of every clustering algorithm is to have as few input parameters as possible, while still delivering excellent results. This assessment dimension focusses on the description of all input parameters an observed algorithm has and thus of all the domain knowledge a potential user of this algorithm has to think of in advance. One of the most important input parameters is the amount of desired clusters k, which is why k as an input parameter is explicitly discussed in a separate assessment dimension.

**Outputs**

The interpretability of clustering results is crucial for solving the initial business problem. Therefore, it is of high importance to provide detailed information regarding the output structure of an observed algorithm to the user. To give an example: k-means and k-medoids algorithms (such as PAM or CLARA) both deliver centers of gravity as their clustering results. While for k-means this center of gravity is a fictional point in the feature space based on the calculation of the mean within the cluster, the output of a k-medoids algorithm is actually a real data point of the cluster that is placed nearest to the fictional center of gravity.

**Size of the dataset**

The size of the dataset has a major effect on the clustering quality and the corresponding processing time. Some clustering algorithms are more accurate and efficient when being applied to small datasets, others when being applied to large ones (Fahad et al. 2014). Algorithms for large-scale data often utilize methods to reduce or aggregate the size of the original dataset such as sampling or applying a grid. It is then a tradeoff between accuracy and processing time. The assessment of this dimension distinguishes between "large" and "small" datasets.

**Handling high dimensional data**

High dimensionality in the data refers to the containment of a large number of features. It is a particular important capability of clustering algorithms to being able to accurately process such data because many applications these days require the analysis of high dimensional data. Pictures for example contain thousands of pixels as features, text data analysis or voice recognition face the same problem. The handling of high dimensional data is often not possible due to the "curse of dimensionality" and many dimensions may not be relevant. With an increasing number of dimensions, the data becomes increasingly sparse. As a result, the distance measurement between data points becomes meaningless and the average density of data points anywhere in the data space is likely to be low (Fahad et al. 2014). An algorithm is rated as eighter being capable of handling high dimensionality or not.

**Handling noisy data and outliers**

In general, real world datasets are not pure and always contain some noise or outliers. This makes it difficult for an algorithm to find suitable clusters. While some algorithms are sensitive to noise and outliers and thus explicitly incorporate them into the cluster building process, other algorithms are able to detect noise as well as outliers and are able to ignore them in the process of finding suitable clusters (Fahad et al. 2014). Again,

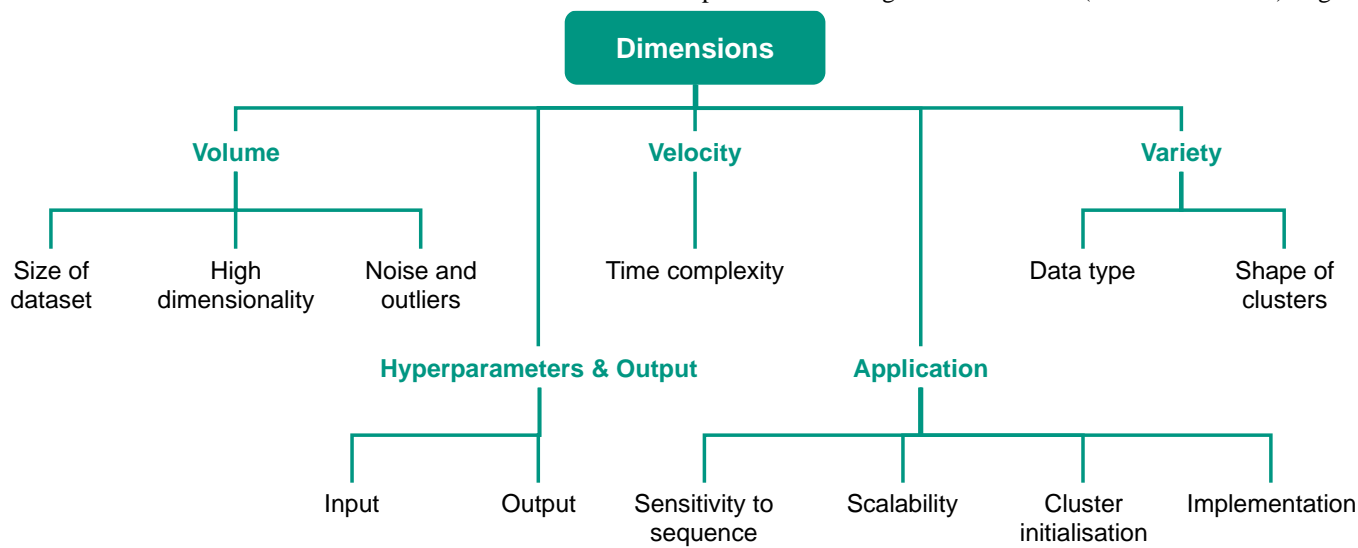

**Fig. 6 Assessment dimensions for the evaluation of clustering algorithms**

an algorithm is assessed as being capable of handling noise and outliers or not.

**Type of dataset**

Most algorithms are designed to be applied on eighter numerical or categorical data. Others are optimized for the application to special data types such as spatial data and only a few algorithms are able to handle combinations of different data types. In the process of selecting a suitable clustering algorithm for a given dataset, it is thus of high importance to choose an algorithm that is able to handle the corresponding data type.

**Shape of clusters**

The way a clustering algorithm processes a dataset in order to find suitable clusters determines the shape of the clusters that can be found. The usage of traditional distance measures for example, generates a clustering result where clusters only have convex shapes and are thus only capable of detecting clusters correctly if they are present in the dataset in a convex manner. Other algorithms are able to detect arbitrary cluster shapes in the data, for instance if the approaches build clusters based on density instead of distance measures.

**Time complexity**

The most crucial factor regarding the time complexity of an algorithm is the size of the dataset, but also other input parameters can influence the processing time such as the number of clusters or the shape of an applied grid on the original data. In the best case, the relationship between processing time and data input is linear or less, but there are some algorithms that are more complex and follow a squared dependency for instance. Users of clustering algorithms have to keep in mind the time complexity of an algorithm together with the database they want to apply it to in order to avoid long processing times, which can get impractical easily (Fahad et al. 2014). In the assessment, time complexity is broken down into "low", "medium" or "high" to keep the overview.

**Sensitivity to sequence of inputting data**

An algorithm is sensitive to the sequence of inputting data, if a different input sequence leads to a different clustering result or different processing times. This assessment criterion is especially important for algorithms that process a dataset incrementally. Incremental clustering is for example used, if the volume of the overall dataset goes beyond the memory capacities of the processing computer. Data then needs to be sorted before processing and an insensitivity to the order is desirable (Sheikholeslami et al.; Xu and Tian 2015). An algorithm can be "insensitive", "moderately insensitive", or "highly sensitive" to the inputting sequence.

**Scalability**

As the size of the dataset, the number of dimensions, or the number of clusters increase, the performance of an algorithm varies in general. Scalability refers to the capability of an algorithm to maintain its performance, no matter how the above-mentioned parameters change. Facing current data clustering problems, algorithms scaling well with increasing samples, dimensions, and clusters are preferable. Scalability is rated as "low", "medium", or "high" in assessment process.

**Cluster initialization**

As already stated above, the practical application of an algorithm depends on the amount and the complexity of corresponding input parameters. The most important among them is the number of clusters. While some algorithms need to define the number of clusters in advance, others do not which is preferable in general. Fixing the number of clusters a priori requires basic domain knowledge. To give an example: clustering the famous iris flower dataset with the number of clusters being fixed to k=2 or k=4 produce useless results. The applicant of the algorithm needs to know in advance that the algorithm should find k=3 clusters in the data, in order to get good results. But in clustering data of high volume and complexity nowadays, the exact number of clusters is often not known a priori, requiring complex and time-consuming approaches in order to find a good number for k before the algorithm is even applied (Fahad et al. 2014; Xu et al. 1998). If the user of an algorithm is therefore not sure about a suitable number of clusters in the given dataset in advance, algorithms that abstract from this input parameter are preferable.

**Implementation**

Even the best-suited algorithm in theory is worth nothing, if there is no good implementation available that can be easily adopted by the average user. For most clustering algorithms, frameworks and libraries for implementation are available in the most important programming languages for data analytics such as R, MATLAB or Python. But there are still some that did not find their way into practical and easy-adoptable usage yet.

### 3.2 ASSESSING CLUSTER VALIDATION

In order to answer RQ2 four dimensions for the assessment of cluster validation concepts are introduced. They describe how the found clusters are used in the validation process and how they affect the results and applicability of the validation index. Another criteria is the capability of handling noise without any pre-processing. Moreover, complexity and computational cost of an index are contemplated. The goal of this analysis is to provide an overview over different validation indices to simplify proper selection for users. All dimensions are visualized in Fig. 7. Furthermore, some additional information about the indices such as interpretation of the values, additional properties or unsolved problems will be provided.

#### 3.2.1 *Definition of the assessment dimensions*
**Capability of detecting arbitrary cluster shapes**

Like clustering algorithms, validation indices have similar problems when it comes to detecting arbitrary cluster shapes and therefore should not be used if such cluster shapes are expected or are likely to be generated by the used algorithm. For example, a clustering solution generated by the DBSCAN algorithm should be validated with an index that can detect arbitrary clusters. We distinct between three different levels of capability – low, medium, and high. Indices with the level low have a hard time finding arbitrary clusters and are mainly used if the cluster are spherical/convex and mostly well separated and compact. Indices with a medium score are indices that work well on non-standard geometry shaped clusters but fail to detect ring shaped or extraordinarily curved shaped clusters e.g. the $S_{Dbw}$-Index. Indices that can find arbitrary cluster shaped belong to the highest level.

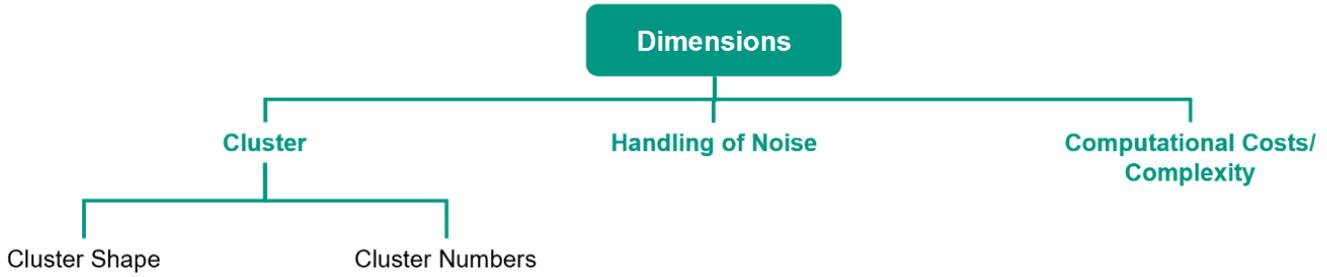

**Fig. 7 Cluster validation dimensions**

**Influence of the number of clusters**

Another way in which the found clusters may influence the results of a validation process is by the number of clusters, since an index might be biased towards a high number of clusters.

**Handling of noise**

As the capability of detecting arbitrary clusters, the importance of this dimension is also dependent on the used algorithm and only comes into play if algorithms like DBSCAN are used that actively detect noise and outliers. It needs to be said, that this dimensions only focuses on whether the index in capable of handling noise without any pre-processing or extension of the index.

**Computational cost and complexity**

As the relevance of big data and especially the pure amount of data steadily increases, it is important to use an index that can evaluate high dimensional at affordable costs. The dimensions computational cost and complexity try to account for that.

*3.2.2    Overview of selected indices*
Here we will give you a short introduction into the indices[4] that we have selected so far.

**Dunn-Index (Halkidi et al. 2002)**

The Dunn-Index rather describes a family of indices than an index itself, where all indices follow the following formula. The concrete indices may differ in the calculation of the inter-cluster[5] and intra-cluster[6] distance.

$$Dunn - Index = \min_{1 \leq i \leq k} \left\{ \min_{1 \leq j \leq k \ und \ i \neq j} \left\{ \frac{d(X_i, X_j)}{\max_{1 \leq c \leq k} \{d(X_c)\}} \right\} \right\}$$

**Average Silhouette Score (Rousseeuw 1987)**

The average Silhouette Score (SilSC) works best, when the distance between the objects is given by a ratio scale (e.g. euclidean distance) and for compact, well separated clusters.

For every object the individual silhouette score $s(i)$ is calculated as followed:

$$s(x_i) = \frac{b(x_i) - a(x_i)}{\max\{a(x_i), b(x_i)\}}, where \ s(x_i) \leq |1|$$

$a(x_i)$ describes the average dissimilarity between all objects $x_i$ in the same cluster and $b(x_i)$ the minimum of the average dissimilarity of $x_i$ to all the object of the other clusters. The average Silhouette Score of the partition C is defined as average over all individual silhouettes scores:

$$S(C) = \frac{1}{|X|} \sum_{i=1}^{n} s(x_i)$$

**SDBW (Vazirgiannis and Halkidi 2001)**

This index calculates its final score by adding a measure of the compactness and separation. The compactness is measured by the intra-cluster-variance and the separation by the inter-cluster-density which are defined as followed:

$$\rho_{intra}(C) = \frac{1}{k*(k-1)} \sum_{j=1}^{k} \left( \sum_{m=1, i \neq j}^{n} \frac{density(u_{ij})}{\max\{density(\bar{c}_j), density(\bar{c}_m)\}} \right)$$

$$\sigma_{intra}(C) = \frac{\left(\frac{1}{k} \sum_{j=1}^{k} \|\sigma(\bar{c}_j)\|\right)}{\|\sigma(X)\|}$$

The density function is a measurement of the number of points in the neighborhood of a given point and further defined as followed:

$$density(u) = \sum_{l=1}^{n_{jm}} f(x_l, u), \ with \ n_{jm} = |X_j \cup X_m|$$

$$f(x, u) = \begin{cases} 0, if \ d(x, u) > stdev^7 \\ 1, sonst \end{cases}$$

---

[4] we will use the term indices if we consider concrete implemented validation metrices
[5] $d(X_i, X_j)$
[6] $d(X_c)$
[7] Stdev is the Standard Deviation

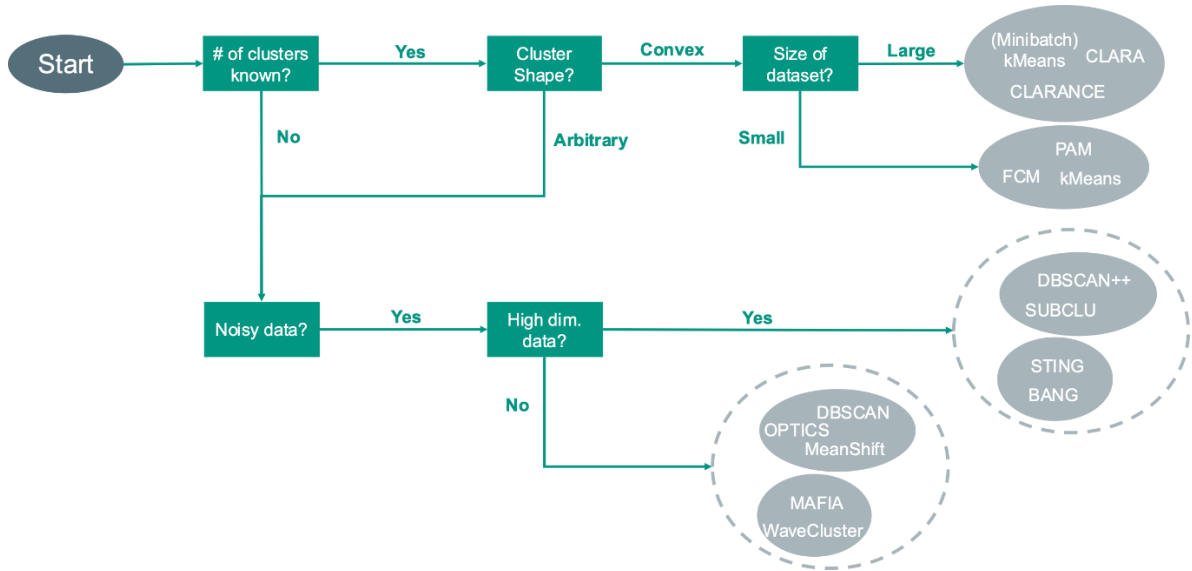

**Fig. 8 Decision tree for algorithm selection**

**CDBW (Halkidi and Vazirgiannis 2008)**

To account for the shape of the clusters in partition C, first representative points are calculated for each cluster. Based on these points the final score is calculated by multiplying a measurement for the separation with respect to compactness ($SC(C)$) and the cohesion of the clusters followed

$$CDbw = SC(C) * Cohesion(C), k > 1$$

The Separation with respect to the compactness is a combination of a measurement for separation and compactness. Where the Cohesion tries to account for change in density within a cluster itself with respect to the overall density within a cluster and is defined as followed:

$$Cohesion(C) = \frac{Compactness(C)}{1 + Intra\_change\ (C)}$$

**DBCV (Moulavi et al. 2014)**

The DBCV-Index that is especially for the evaluation of density-based clustering and is there for rather defined by the means of density than by distances. At first the all-point-core-distance is calculated which equals "the invers of the density of each point with respect to all other points in its cluster". This is used to "defined a symmetric reachability distance which is then employed to build a Minimum Spanning Tree (MST) inside each cluster" which captures shape as well as density of the cluster. Based on this MST the density sparseness of a cluster (DSC) and the density separation of a pair of clusters (DSCP) in extracted, where DSC is defined as the "maximum edge weight of the internal edges of the MST of the cluster and DSPC as the "minimum reachability distance between the internal nodes of the MST of the respective clusters. Out of

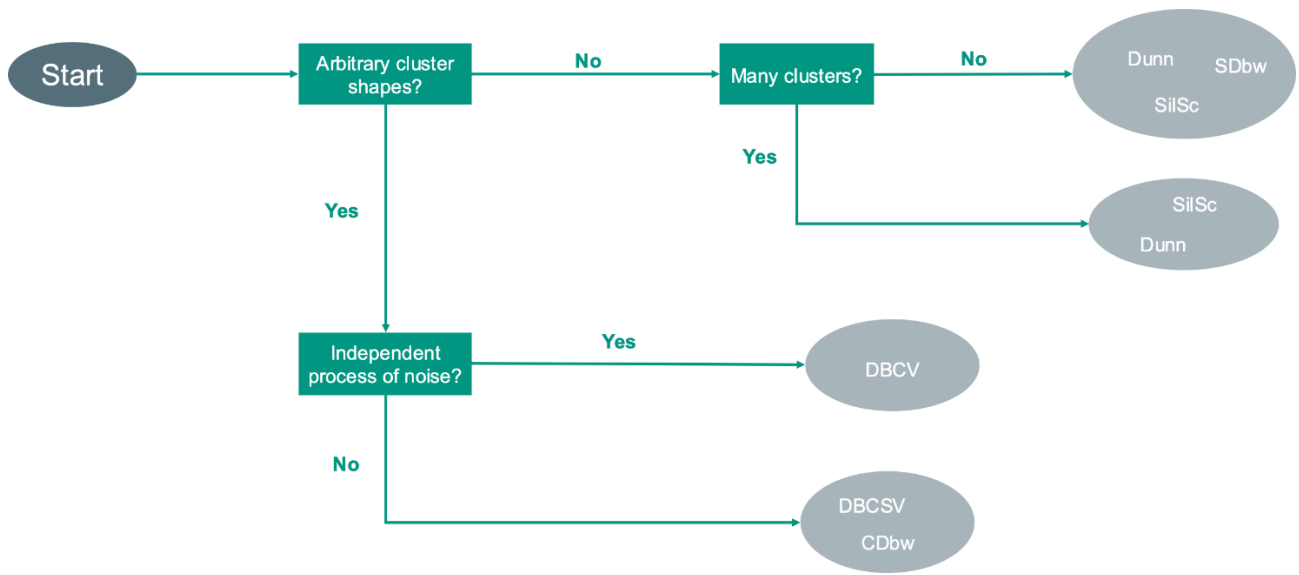

**Fig. 9 Decision tree for validation**

those the validity index of a Cluster $V_C$ is computed. The final index is defined as followed:

$$DBCV\ (C) = \sum_{(i=1)}^{l} \frac{|C_i|}{|O|} V_C(C_i)$$

### 3.3 CLUSTERING CHEAT SHEET

In this chapter, the assessment dimensions from 3.1 and 3.2 are used for the evaluation of overall 34 clustering algorithms and 9 validation indices.

We provide a repository hosted at a gitlab repository for extension of categorized cluster algorithms and evaluation metrics (https://git.scc.kit.edu/ml-wzmm_public/a-clustering-algorithm-review). An interactive web-based version can be found on https://jhillenbrand.github.io/cluster_table.html.

Starting with a clustering problem and the corresponding data set, this holistic comparison table can be used as a systematic selection logic for finding suitable clustering algorithms and validation concepts. A user only has to filter the columns for suitable descriptions. The visualized decision trees exemplary show how the procedure could look like from the user's perspective. To make the selection logic even more convenient, chapter four provides some pseudocode that extracts necessary domain knowledge regarding the data characteristics automatically and directly from the raw data set.

As Fig. 8 demonstrates, a user would in the first step determine whether the number of clusters to be detected in the data set is known in advance. Together with the assumption that the clusters to be detected are of convex shape, the user again follows the upper path. Those two decisions already lead to the recommendation of using a partitioning-based clustering algorithm. But the range of different algorithms in that group is still large. So in the next node, a user defines if the data set is comparatively small (see chapter four for more insights on what that could mean) or large. For small data sets, the procedure recommends using classical partitioning-based algorithms such as k-means or PAM. For larger data sets, according to the insights from the comparison table, some variants of PAM are more useful, since they can be applied to sub samples for a better runtime, namely CLARA or CLARANCE.

Regarding the selection of a suitable validation concept, Fig. 9 serves as an example. A user derives from domain knowledge that clusters may take on arbitrary shapes and assumes that the

**Routine 1:** *Compute the operations per sample for each algorithm*

| | | |
|---|---|---|
| 1. | **function** *RankComputingVelocity*(dataSet) | |
| 2. | $(n, m) \leftarrow Size$(dataSet) | |
| 3. | steps ← List.Empty() | # create empty list |
| 4. | algorithms ← GetAlgorithmList() | # function retrieves all algorithms |
| 5. | | # available for investigation |
| 6. | **for** $k \leftarrow \{2, 100, 1000\}$ | |
| 7. | **for** algorithm **in** algorithms | |
| 8. | $O(n, m, k) \leftarrow$ algorithm.TimeComplexity | # stores the function |
| 9. | | # handle of the time |
| 10. | | # complexity of the algo- |
| 11. | | # rithm |
| 12. | computingStepsPerSample ← $O(n, m, k) / n$ | |
| 13. | steps.Add(computingStepsPerSample) | |
| 14. | **end for** | |
| 15. | steps.Sort(ASCENDING) | |
| 16. | Print(steps) | # print the computing steps for each |
| 17. | | # algorithm for k |
| 18. | steps ← List.Empty() | |
| 19. | **end for** | |
| 20. | **end function** | |

***Routine 2:*** *Find the dimension category for a dataset*

| | |
|---|---|
| 1. | **function** dimensionCategory = *GetDimension*(dataSet) |
| 2. | $(n, m) \leftarrow$ Size(dataSet) |
| 3. | **if** $m \leq 10$ |
| 4. | dimensionCategory $\leftarrow$ LOW |
| 5. | **else** |
| 6. | dimensionCategory $\leftarrow$ HIGH |
| 7. | **end if** |
| 8. | **return** dimensionCategory |
| 9. | **end function** |

data set is rather noisy. The first point –arbitrary shapes – leads the user along the lower branch. Next the user decides that there should not be an additional preprocessing step in order to deal with the noise in the data. So, the cheat-sheet will recommend the DBCV as a possible validation index since it can deal with arbitrary cluster shapes and with noise.

## 4 ALGORITHM SELECTION ASSISTANCE

Following our assessment dimensions in chapter 3.1, we provide pseudo code to assist the selection of a corresponding clustering algorithm based on sample input data from the contemplated dataset. These approaches assume no general applicability, but performed well as sophisticated guess on arbitrary datasets.

Routine 1 expects a raw data set as input and extracts the number of samples $n$ as well as the dimensionality $m$. For a varying number of cluster $k \in \{2, 100, 1000\}$, and for each algorithm from the collected table of clustering algorithms, the function then computes the necessary operations per sample for performing a clustering task with the above-mentioned parameters. Therefore, the big-O notation $O(n, m, k)$ from the table is used. The function then returns the computational steps per sample in ascending order for each algorithm, so that users can select the most efficient clustering algorithm for the given data set.

Routine 2 expects a raw data set as input. It extracts the information about the dimensions of the dataset where $n$ contributes to the number of observations and $m$ to the number of different features. A logic is implemented that returns a label for the dimension category based on $m$. This should help the user to identify whether he deals with low or high dimensional data.

Routine 3 expects a raw data set as input. It extracts the information about the dimensions of the dataset where $n$ contributes to the number of observations and $m$ to the number of different features. A logic is implemented that returns a label for the size category based on $n$. This should help the user to

***Routine 3:*** *Find the size category for a data set*

| | |
|---|---|
| 1. | **function** sizeCategory = *GetSize*(dataSet) |
| 2. | $(n, m) \leftarrow$ Size(dataSet) |
| 3. | **if** $n \leq 50$ |
| 4. | sizeCategory $\leftarrow$ SMALL |
| 5. | **else if** $n \leq 10.000$ |
| 6. | sizeCategory $\leftarrow$ MEDIUM |
| 7. | **else** |
| 8. | sizeCategory $\leftarrow$ LARGE |
| 9. | **end if** |
| 10. | **return** sizeCategory |
| 11. | **end function** |
| **Routine 4:** *Assessment of noise in a data set* | |
| 1. | **function** output = *noiseAssessment*(DataSet) |
| 2. | DataSetNormalized ← z-score(DataSet) |
| 3. | DataSetNoise ← addNoise(DataSetNormaliszed) |
| 4. | Dist2Mean ← getDistanceToMean(DataSetNormalisized) |
| 5. | Dist2MeanNoise ← getDistanceToMean(DataSetNoise) |
| 6. | Hist_DistToMean ← histogramm(Dist2Mean) |
| 7. | Hist_DistToMeanNoise ← histogramm(Dist2MeanNoise) |
| 8. | Plot Figure |
| 9. | **end function** |
| 10. | |
| 11. | **function** output = *addNoise*(DataSet) |
| 12. | $(n, m)$ ← $dim(DataSet)$ |
| 13. | $a$ ← $min(DataSet)$         # per dimension |
| 14. | $b$ ← $max(DataSet)$         # per dimension |
| 15. | $Noise$ ← $(b - a) * rand(n, m) + a$     # rand draws from uniform distribution |
| 16. | $DataSet$ ← $DataSet.join(Noise)$ |
| 17. | *return DataSet* |
| 18. | **end function** |
| 19. | |
| 20. | **function** output = *getDistanceToMean*(DataSet) |
| 21. | $mean$ ← $getMean(DataSet)$ |
| 22. | $distMean$ ← $dist(mean, DataSet)$ *calculate Euclidean distance* |
| 23. | $distMean$ ← $sorted(distMean, ascending)$ |
| 24. | *return distMean* |
| 25. | **end function** |

identify whether he deals with with a small, medium, or large dataset. The logic is based on (Pedregosa et al. 2011).

Routine 4 tries to extract information about whether the dataset contains a lot of noise. The raw data set is used as input parameter and standarsizes using z-score standarization. Artifical noise, which is simulated by drawing from a continous uniform distribution for each dimension, is added to the Dataset and stored seperately. At last, for both data sets the euclidean distance to the mean is calculated, sorted ascendantly and plotted. The generall idea is that intoduction of noise to a already noisy data set should have less impact on the distribution of the distance to mean value than on a dataset that does not cointain noise.

Of course this method is rather superfical and works under hard assumptions and limitation and should be understood as a first example in order to help an unexperienced user to gain knowledge about noise in the data. The corresponding plots are displayed in Fig. 10 and Fig. 11.

In order to identify, if the underlying data and its corresponding clusters adhere to convex shapes, we propose a similarity comparison of the actual cluster boundary points with a fitted ellipsoid. By definition ellipsoidal volumes are convex and can therefore represent a convex data distribution as approximation. Hence, we compute the total volume $V_E$ encompassed by an ellipsoid fitted to the cluster points and form the ratio with the actual volume $V_B$ encompassed by the boundary points of the cluster. If this ratio $V_E/V_B$ subceeds an arbitrary threshold (here we use $\tau = 0.7$), the underlying cluster points cannot be well fitted by an ellipsoid. Then, we suggest using an algorithm, that is capable of handling non-convex cluster shapes.

| | **Routine 5a:** *Estimate shape of clusters* |
|---|---|
| 1. | **function** isConvex = estimateConvexity(clusterPoints) |
| 2. | boundaryPoints = GetBoundaryPoints(clusterPoints)   # retrieve boundary points |
| 3. | # of clusterPoints using |
| 4. | # alpha shapes |
| 5. | $V_B \leftarrow$ ComputeBoundaryVolume(boundaryPoints)   #using convex hull |
| 6. | $V_E \leftarrow$ ComputeEllipsoidVolume(boundaryPoints)   # using Khachiyan Algorithm |
| 7. | **if** $V_E / V_B > 0.7$ |
| 8. | isConvex $\leftarrow$ **true** |
| 9. | **else** |
| 10. | isConvex $\leftarrow$ **false** |
| 11. | **end if** |
| 12. | **return** isConvex |
| 13. | **end function** |

The pseudo code for routines 5a and 5b illustrates the procedure on how we detect the convexity of cluster points. Routine 5a assumes the user can already supply the points of a separated cluster to our method. It then starts with retrieving the boundary points of this cluster. In our implementation example we are using MATLAB to find the boundary points using alpha shapes (MathWorks 2014). Next follows the computation of volumes for the boundary points using the convex hull method (MathWorks 2006) and the ellipsoid using Khachiyan's algorithm (Moshtagh and Nima 2009). If the ratio of these volumes subceeds a defined threshold, the cluster points are labeled non-convex and an appropiate cluster algorithm can be selected from the provided collection.

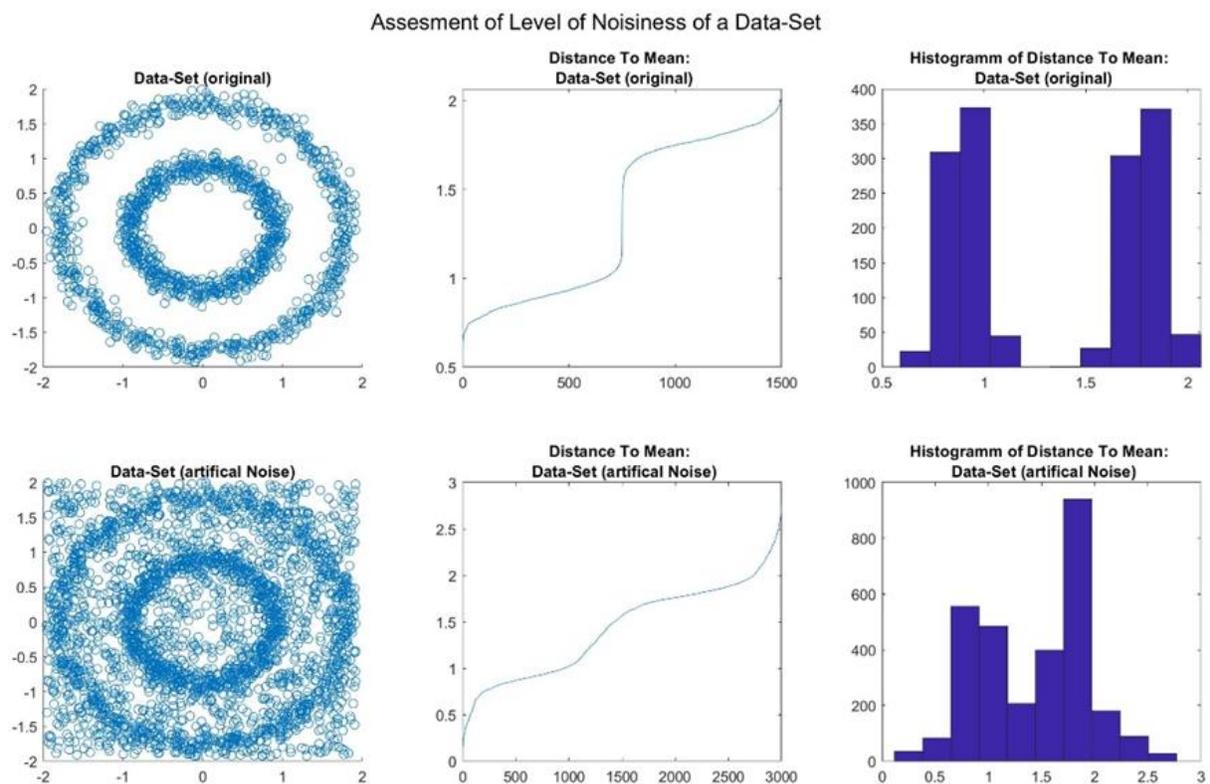

**Fig. 10 Plot of noise assessment function for an already noisy data set**

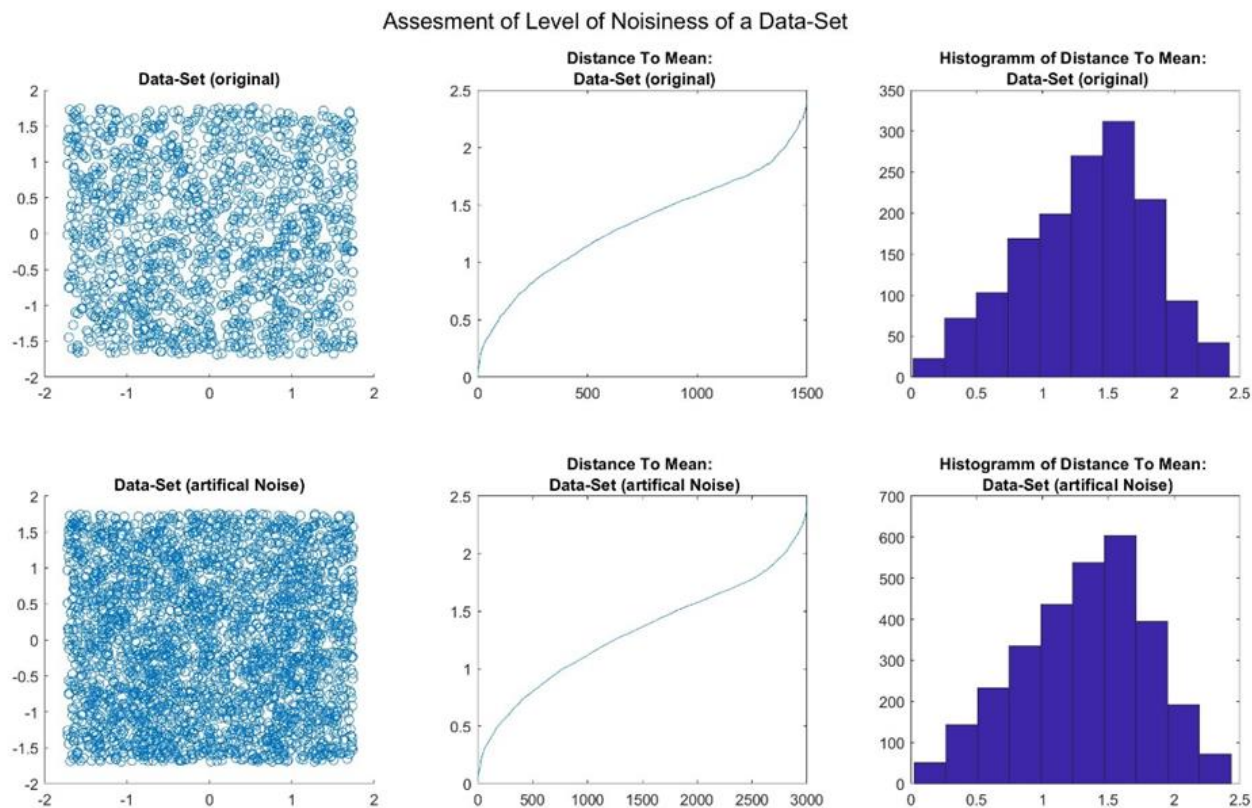

**Fig. 11 Plot of noise assessment function for an arbitrary data set**

With Routine 5b the problem is addressed in a different manner: Here we assume the user can specify the number of clusters $k$ in his dataset (or a subset) a-priori. We then use k-means to find the cluster groups and compute volumes of border points and ellipsoids according to 5a. Then, based on the threshold $\tau$ can also be specified, if these cluster groups adhere to an ellipsoidal shape. These procedures rely of course on the definition of threshold $\tau$, but deliver a geometrically motivated and comparable relative measure for the convexity of cluster shapes.

## 5 DISCUSSION

In this chapter, the systematic proposal from chapter three and four is evaluated in regard to the fulfillment of the initial goal statement. To emphasize the objective once again, this work aims for identification of a systematic logic for the selection of suitable clustering algorithms and a corresponding validation concept for a given dataset and business problem. The two research questions derived from this statement were (1) the identification of appropriate assessment dimensions regarding the evaluation of clustering algorithms to a certain data set and business problem and (2) the identification of appropriate assessment dimensions regarding the selection of a suitable clustering validation concept.

**Cluster Algorithms**

In this proposal, twelve assessment dimensions are introduced to evaluate clustering algorithms. As shown in Fig. 6, the dimensions can be grouped into five categories, each highlighting the applicability of clustering algorithms from a different perspective. The categories "Volume", "Velocity" and "Variety" serve as a big data perspective. The category "Hyperparameters & Output" ensures that domain knowledge is considered in the proposal and the category "Application" ensures that the whole process is executed in an application-oriented way. With the help of the holistic comparative study in chapter 3.3 and the algorithms introduced in chapter four, the assessment dimensions serve as a decision support tool for users and thus fulfill research question 1.

Besides the fulfillment of the subject of this paper, there are some properties that need to be discussed in more detail: The first issue concerns the expression of the different assessment dimensions for the clustering algorithms. This problem occurs independently from the dimensions the assessment is based on and lies in the nature of the resources used for evaluation, which have their origin in different times. (Huang 1997) for example defines the k-modes algorithm as a clustering approach for very large datasets in 1997, while today the understanding of the term "very large" probably describes something completely different. Still, there is sometimes no other resource available that investigates the application of k-modes to large data based on today's understanding of "large". Another issue is that assessments of different algorithms are executed independently from each other. This is closely related to the previous issue and

| | |
|---|---|
| *Routine 5b:* *Estimate shape of clusters* | |
| 1. | **function** isConvex = *estimateConvexity*(dataSet, $k$) |
| 2. | clusters ← KMeans(dataSet, $k$)     # apply k-means on data set |
| 3. | volumeRatios ← List.Empty() |
| 4. | **for** cluster **in** clusters |
| 5. | boundaryPoints = GetBoundaryPoints(clusters.clusterPoints) |
| 6. | $V_B$ ← ComputeBoundaryVolume(boundaryPoints) |
| 7. | $V_E$ ← ComputeEllipsoidVolume(boundaryPoints) |
| 8. | volumeRations ← volumeRations + $V_E / V_B$ |
| 9. | **end for** |
| 10. | **if** volumeRatios / $k$ > $\tau$ |
| 11. | isConvex ← **true** |
| 12. | **else** |
| 13. | isConvex ← **false** |
| 14. | **end if** |
| 15. | **return** isConvex |
| 16. | **end function** |

is based on the problem that there is no general understanding of terms defined throughout the different resources investigating clustering algorithms. While one author defines "high-dimensional" as data with a feature space larger than 5, another one may define "high-dimensional" as data with a feature space larger than 30. Mostly, there is simply no detailed discussion on what is actually meant by the terms used in the resources. A further problem that is related to the assessment procedure of algorithms in general is that the same algorithm, evaluated based on the same criteria, is evaluated differently when observing different resources. For instance: The time complexity of COBWEB is $O(n^2)$ according to (Fahad et al. 2014), $O(n)$ based on (Berkhin 2006), and "low" according to (Xu and Tian 2015) which in general stands for a linear dependency. Actually, the dimension "time complexity" of an algorithm does not allow much room for interpretation, but still there seems to be different assessment results. The last issue in regard to the assessment procedure is that the evaluation of an algorithm might be biased. This is especially the case if the authors of an algorithm itself rate its performance. Last, some differences between clustering algorithms are not generally applicable to all algorithms, so introducing a general assessment dimension is not useful. As a result, some algorithms look like being applicable to the same kind of problems although they follow a different goal. To give an example: DBSCAN and SNN are rated similar in all dimensions, but SNN is best-suited for finding cluster of different densities, while DBSCAN is not. Defining an assessment dimension that focusses on the detection of clusters of different densities is not target-oriented, since partitioning-based algorithm have no use for this. Instead, such approaches could be applied on a more detailed level, for example the evaluation of density-based clustering methods or, as already carried out by (Berkhin 2006), the evaluation of grid-based algorithms only.

**Cluster Validation**

As stated above the selection of an appropriate validation index depends on properties of the data as well as the used clustering algorithm. The literature proposes a lot of different indices, so selecting a suited one isn't a trivial task. In order to support researchers and users by selecting an index, we proposed the assessment dimensions cluster shapes, cluster numbers, handling of noise and complexity/computational costs, which we consider to be the most important. We categorized the mentioned indices exemplarily. The next step will be using those dimensions to further classify other indices and, if necessary, add more assessment dimensions to the list. Additionally, an empirical study analyzing the benefits of the applying the proposed logic especially for unexperienced user would underline the necessity for the proposed taxonomy. Furthermore, the assessment dimensions might be used by researchers proposing new validation indices in order to analyze their proposal and make sure it can be easily applied by a wide range of users. Also, additional research is needed to support user not only with the categorizations over assessment dimension but also within e.g., whether the Dataset is likely to contain mostly convex clusters. At last, since the selection of a validation index is also influenced by the used algorithm, a next step would be to propose a set of validation indices that are appropriate for the algorithm and data.

# 6 CONCLUSION

To address RQ1, twelve assessment dimensions for the evaluation of clustering algorithms were presented in chapter 3.1. The procedure could successfully be applied to 34 traditional and novel clustering algorithms. As a result, the holistic comparison of clustering algorithms was presented in form of a table that rates each single clustering algorithm based on all the assessment criteria.

The table is accessible for the public on a GitLab repository (https://jhillenbrand.github.io/cluster_table.html) and can be utilized by users in order to select an algorithm that perfectly matches the users' underlying data clustering problem. The extraction of the assessment dimensions from a given data set can be difficult sometimes. For this purpose, chapter four describes different functions that can be applied to the data set of interest. As a result, users can directly derive a list of suitable algorithms from collection of available algorithms that are suitable for the contemplated clustering problem.

The validation of the clustering results is an important step, when extracting knowledge from data using unsupervised methods. In order to satisfy RQ2 of helping users throughout the entire process and not only the clustering part itself, analogous to the assessment dimensions for clustering algorithms, dimensions for validation indices were proposed as well. The paper gave a short introduction into cluster validation research and describes the main categories of validation methods. A short example why the selection is not trivial and depends on data and algorithm was given. Furthermore, the general properties of validation methods were described. Then four assessment dimensions were proposed, defined and some exemplarily indices were described.

## 6.1 OUTLOOK

First of all, the procedure and the usefulness of the assessment dimensions for identifying suitable algorithms need to be proven in practice. A suitable approach for validation might be the comparison of clustering results for different datasets that are on the one hand clustered by using a random clustering algorithm and on the other hand by using clustering algorithms that are recommended based on the comparative assessment dimensions from chapter three. The presented code in chapter four will serve as a basis for doing so. The results to be expected are that clustering algorithms selected based on the logic presented in this work perform significantly better than randomly chosen algorithms.

In chapter four, the proposed method for assessing the noisiness in a dataset was only tested on artificial datasets. As a next step known real world data sets should be tested as well. Additionally, at the moment the method only introduces uniformly distributed noise in the data set. A possible adoption would be to introduce noise that is based on kernel density estimation of the dataset in order to create more realistic noise. Another aspect could be to not introduce the noise on the level of the actual data distribution but on the distribution of the distance to mean values to avoid operation a lot in high dimensional space.

Chapter five pointed out some problems related to the fact that an assessment is in general carried out by using different resources from different authors and different times. Those issues should be faced in further research, for example by performing a holistic evaluation of clustering algorithm similar to the presented approach, but not based on reviews of existing literature. Instead, the assessment could be carried out by relying on own results that support a common understanding of the single assessment criteria throughout all observed algorithms.

Last, the presented approach can always benefit from introducing new assessment dimensions or apply it to a wider range of clustering algorithms and validation concepts in future research.